\documentclass[10pt, a4paper]{article}

\pdfoutput=1 

\usepackage{lrec}
\usepackage{multibib}
\newcites{languageresource}{Language Resources}
\usepackage{graphicx}
\usepackage{tabularx}
\usepackage{soul}
\usepackage{float}

\usepackage{flushend}

\usepackage{epstopdf}
\usepackage[utf8]{inputenc}

\usepackage{hyperref}
\usepackage{xstring}

\usepackage{color}

\title{English-Catalan Neural Machine Translation in the Biomedical Domain through the cascade approach}

\name{Marta R. Costa-jussà, Noe Casas and Maite Melero$^*$}

\address{TALP Research Center - Universitat Polit\`ecnica de Catalunya, Barcelona \\
         $^*$ OTG Plan de Tecnologías del Lenguaje \\
         marta.ruiz@upc.edu; contact@noecasas.com; maite.melero@upf.edu\\
}

\abstract{This paper describes the methodology followed to build a neural machine translation system in the biomedical domain for the English-Catalan language pair. This task can be considered a low-resourced task from the point of view of the domain and the language pair. To face this task, this paper reports experiments on a cascade pivot strategy through Spanish for the neural machine translation using the English-Spanish SCIELO and Spanish-Catalan El Peri\'odico database. To test the final performance of the system, we have created a new test data set for English-Catalan in the biomedical domain which is freely available on request.
  \\ \newline  }

\begin{document}

\maketitleabstract

\section{Introduction}

Neural machine translation \cite{sutskever:2014,cho:2014} has recently emerged as a stronger alternative to standard statistical paradigm \cite{koehn:2003}. Among other advantages, neural MT offers an end-to-end paradigm which seems to be able to generalize better from data \cite{bentivogli:2017}. However, deep learning techniques face serious difficulties for learning when having limited or low resources and machine translation is not an exception \cite{koehn:2017}.


English has become the \textit{de facto} universal language of communication around the world. In Catalonia, out of 7.5 million population only around 30\% of people have knowledge of English in all competences\footnote{Data taken from https://www.idescat.cat/}. Therefore, there are many situations where professional or automatic translations are still necessary. One of them is in medical communication patient-physician at the level of primary health care. Also in the biomedical domain it is worth mentioning that Catalonia has become a hub of global biomedical research as proven by the nearly 1\% of global scientific production, 9,000 innovative companies or the fact that the sector raised a record of 153 million of euros in 2016 \footnote{http://cataloniabio.org/ca/publicacions}. Therefore, English-Catalan translation in the biomedical domain is of interest not only in health communication but to properly disseminate the work in such a relevant area for Catalan economy.

English-Catalan in general \textemdash and even more in a closed domain as the biomedical one\textemdash can be considered to be a limited resourced language pair. However, there are quite large amount of resources for English-Spanish and Spanish-Catalan language pairs. Therefore, English-Catalan could take advantage of them by using the popular pivot strategies which consist in using one intermediate language to perform the translation between two other languages.


Pivot strategies have been shown to provide a good performance within the phrase-based framework \cite{costajussa:2012} and also for the particular case of English-Catalan \cite{gispert:2006}. While in the phrase-based context, pivot strategies have been widely exploited, this is not the case for the neural approaches. Pivot studies are limited to \cite{cheng:2017} which considers a single direct approach (the cascade) contrasted with a joint trained model from source-to-pivot and pivot-to-source. Other alternatives when having no parallel data for a language pair are the multilingual approximations where several language pairs are trained together and the system is able to learn non-resourced pairs \cite{wu:2016}.


Another related research area for this study is precisely training translation systems domain-specific tasks, where there are scarce in-domain translation resources. A common approach in these cases consists in training a system with a generic corpus and then, use a small in-domain corpus to adapt the system to that particular domain. In this direction, there is a huge amount of research in the statistical approach \cite{costajussa:2015} and also starting in the neural approach \cite{chu:2017}. Finally, there is am emerging line of research in the topic of unsupervised neural MT \cite{lample18,artetxe18}.

This study designs and details an experiment for testing the standard cascade pivot architecture which has been employed in standard statistical machine translation \cite{costajussa:2012}.

The system that we propose builds on top of one of the latest neural MT architectures called the Transformer \cite{vaswani:2017}. This architecture is an encoder-decoder structure which uses attention-based mechanisms as an alternative to recurrent neural networks proposed in initial architectures \cite{sutskever:2014,cho:2014}. This new architecture has been proven more efficient and better than all previous proposed so far\cite{vaswani:2017}. 


\begin{table*}[h]
\caption{Size of the parallel training corpora}
\label{tab:corpora}
\small
\centering
\medskip
   \begin{tabular}{l rrrrr}
Language Pair &  {\bf Corpus} & Language  & {Segments} & { Words} & {Vocab}\\ \hline
&& En   &  &  $~20.5 \cdot 10^6$ & $296 \cdot 10^3$ \\ 
\raisebox{1.5 ex}[0pt]{En-Es} &  \raisebox{1.5 ex}[0pt]{Biomedical}
& Es  & \raisebox{1.5 ex}[0pt]{$~0.9 \cdot 10^6$}  & $~21.9 \cdot 10^6$ & $309 \cdot 10^3$ \\
\hline
&& Es   & &  $~165.1 \cdot 10^6$ & $ 736 \cdot 10^3$ \\ 
\raisebox{1.5 ex}[0pt]{Es-Ca} & \raisebox{1.5 ex}[0pt]{El Peri\'odico}
& Ca & \raisebox{1.5 ex}[0pt]{$~6.5 \cdot 10^6$} & $~178.9 \cdot 10^6$ & $713 \cdot 10^3$ \\ \hline
\end{tabular}
\end{table*}

\begin{table*}[h]
\caption{Size of the test set}
\label{tab:sets}
\small
\centering
\medskip
 \begin{tabular}{l rrrrr}
Language Pair &  {\bf Corpus} & Language  & {Segments} & { Words} & {Vocab}\\ \hline
&& En &   & $26.1 \cdot 10^3$ & $6.1 \cdot 10^3$ \\
\raisebox{1.5 ex}[0pt]{En-Es} & \raisebox{1.5 ex}[0pt]{Biomedical}
& Es & \raisebox{1.5 ex}[0pt]{1000} & $27.4 \cdot 10^3$ & $6.6 \cdot 10^3$ \\
\hline
&&Es &   & $56.0 \cdot 10^3$ & $12.2 \cdot 10^3$ \\
\raisebox{1.5 ex}[0pt]{Es-Ca} & \raisebox{1.5 ex}[0pt]{El Peri\'odico}
&Ca & \raisebox{1.5 ex}[0pt]{2244} & $60.7 \cdot 10^3$ & $11.7 \cdot 10^3$\\
\hline
\end{tabular}
\end{table*}

\section{Neural MT Approach}

This section provides a brief high-level explanation of the neural MT approach that we are using as a baseline system, which is one of the strongest systems presented recently \cite{vaswani:2017}, as well as a glance of its differences with other popular neural machine translation architectures.

Sequence-to-sequence recurrent models \cite{sutskever:2014,cho:2014} 
have been the standard approach for neural machine translation,
especially since the incorporation of attention mechanisms
\cite{bahdanau:2014,luong:2015}, which enables the system to
learn to identify the information which is relevant for producing each word in the translation.
Convolutional networks \cite{gehring:2017} were the second
\textit{paradigm} to effectively approach sequence transduction
tasks like machine translation.

In this paper we make use of the third \textit{paradigm} for
neural machine translation, proposed in \cite{vaswani:2017},
namely the Transformer architecture, which is based on a
feed-forward encoder-decoder scheme with attention mechanisms. 
The type of attention mechanism used in the system, referred to as
\textit{multi-head attention}, allows to train several attention 
modules in parallel, combining also self-attention with standard 
attention. Self-attention differs from standard attention in the
use of the same sentence as input and trains over it allowing to solve issues as coreference resolution.
Equations and details about the transformer system can be found in 
the original paper \cite{vaswani:2017} and are out of the scope of 
this paper.

For the definition of the vocabulary to be used as input for
the neural network, we used the sub-word mechanism from
\texttt{tensor2tensor} package, which is similar to
Byte-Pair Encoding (BPE) from \cite{sennrich:2016}.
For the English-Spanish language pair, two separate 32K
sub-word vocabularies where extracted, while for
Spanish-Catalan we extracted a single shared 32K sub-word vocabulary for both languages.

\section{Pivot Cascade Approach}

\begin{figure}[h]
\centering
\includegraphics[width=0.7\linewidth]{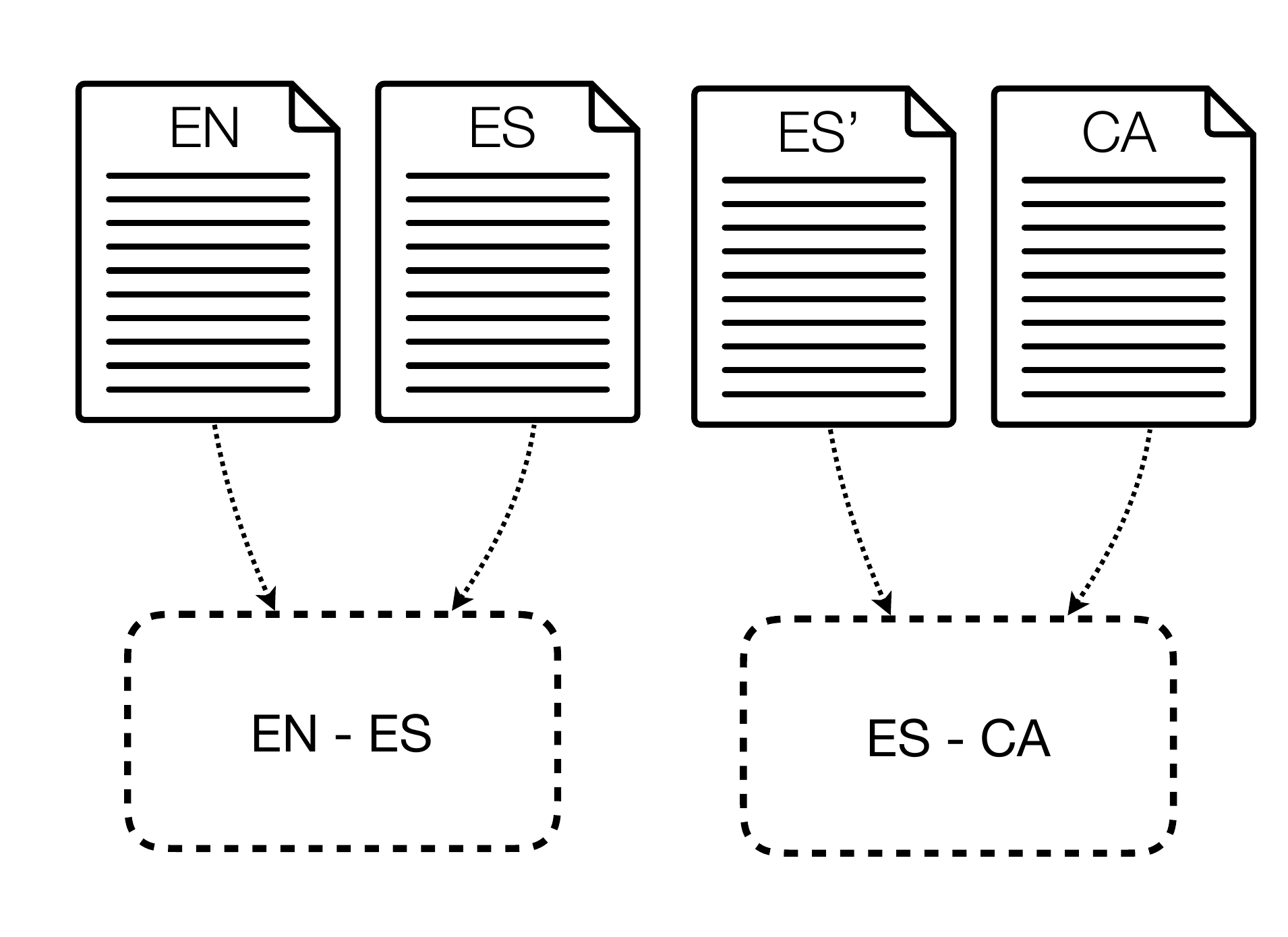}
\caption{Pivot translation systems training.}
\label{fig:pivot}
\end{figure}

\begin{figure}[h]
\centering
\includegraphics[width=1.\linewidth]{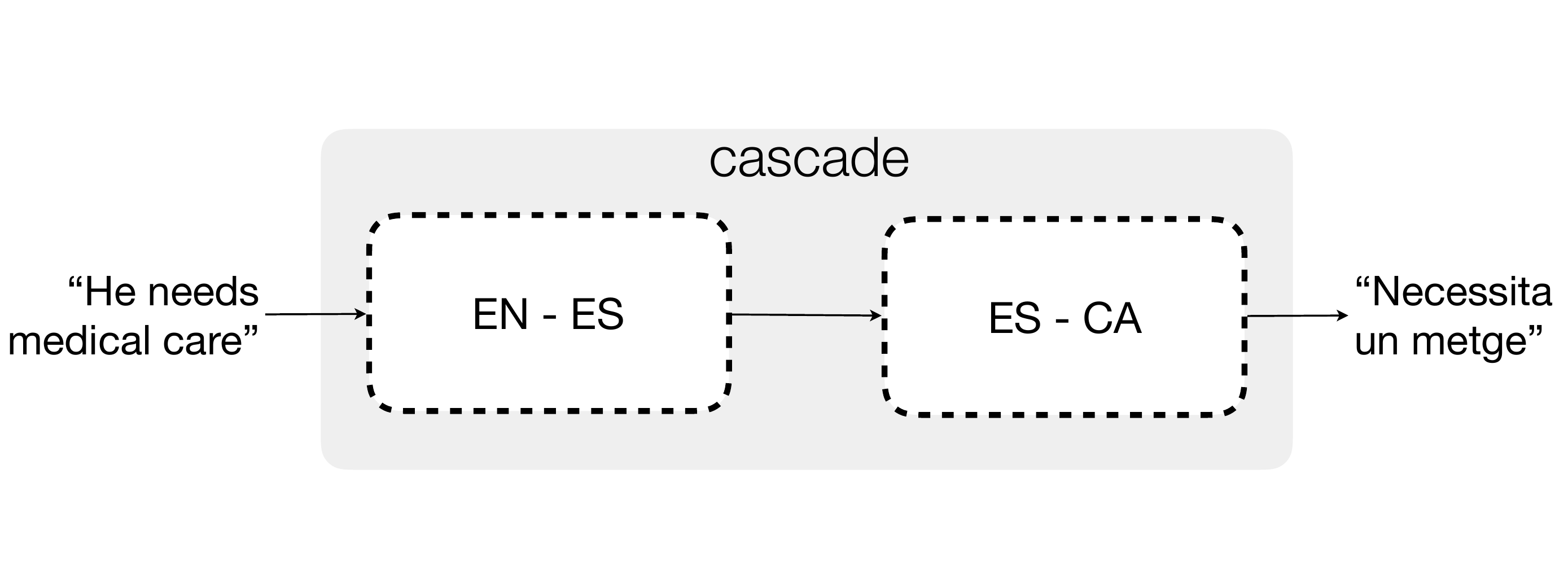}
\caption{Pivot cascading approach at inference time.}
\label{fig:cascade}
\end{figure}

Standard approaches for making use of pivot language translation 
in phrase-based systems include the translation cascade.
The cascade approach consists in building two translation systems: 
source-to-pivot and pivot-to-target. In test time, the cascade 
approach requires two translations. Figure \ref{fig:pivot} depicts
the training of the pivot systems while
figure \ref{fig:cascade} shows how they are
combined to devise the final one. 

\section{Experimental Framework: data resources}

This section reports details on data to be employed in the experiments.

Experiments will be performed for the English-Catalan language pair on the biomedical domain. Resources are the SCIELO database from the WMT 2016 International Evaluation Campaign \cite{wmt}. The English-Spanish corpus is the compilation of the corpora assigned for the WMT 2016 biomedical shared task, gathered from the Scielo  database. The Spanish-Catalan corpus is extracted  from  ten  years
of the paper edition of a bilingual Catalan newspaper,  El  Peri\'odico  \cite{costajussa:2014}. The Spanish-Catalan corpus is partially available via ELDA (Evaluations and Language Resources Distribution Agency) in catalog number ELRA-W0053.

\begin{table*}[t]
\caption{Sample end-to-end English-Catalan translations.}
\label{tab:samples}
\small
\centering
\medskip
\begin{tabular}{p{7cm} p{7cm}}
English & Catalan \\ \hline
\texttt{crustacean diversity and population peaks were
within the range of examples found in worldwide literature .}
&
\texttt{la diversitat de crustacis i els pics poblacionals van estar dins del rang d ’ exemples trobat en la literatura mundial .}\\
\vspace{1mm}\\
\texttt{multivariate analysis and Post-Hoc Bonferroni tests
were used and relative risk and attributable fraction were 
calculated .}
&
\texttt{es va utilitzar anàlisis multivariat i post-Hoc de Bonferroni i es va calcular el risc relatiu i la fracció atribuïble .}
\\
\vspace{1mm}\\
\texttt{this qualitative study used semi-structured interviews , with eight coordinators of the Tuberculosis Control Program in six cities of the state of Paraíba .}
&
\texttt{es tracta d ’ un estudi qualitatiu amb entrevistes semiestructurades , amb vuit coordinadors del Programa de Control de la Tuberculosi en sis municipis de l ’ estat de Paraíba .}
\\
\vspace{1mm}\\
\texttt{there is no statistically significant difference in global and event free survival between the two groups .}
&
\texttt{no hi ha diferència estadísticament significativa en la supervivència global i lliure d ’ esdeveniments entre els dos grups .}
\\
\vspace{1mm}\\
\hline

\end{tabular}
\end{table*}

The size of the corpora is summarised in Table \ref{tab:corpora}.  The corpora has been pre-processed with a standard  pipeline for  Catalan,  Spanish  and  English:  tokenizing and keeping parallel sentences between 1 and 50  words.    Additionally,  for  English and Spanish  we used Freeling \cite{padro:2012} to tokenize pronouns from verbs (i.e. \textit{pregunt\'andose} to \textit{preguntando} + \textit{se}), we also split prepositions and articles, i.e. \textit{del} to \textit{de} + \textit{el} and \textit{al} to \textit{a + el}.  This was done for similarity to English. For Spanish and Catalan, we used Freeling to tokenize the text but no split with pronouns, prepositions or articles was done. The test sets come from WMT 2016 biomedical shared task in the case of English and Spanish. Since we required a gold standard in English-Catalan, we translated the Spanish test set from WMT 2016 biomedical shared task into Catalan. The translation was performed in two steps: we did a first automatic translation from Spanish to Catalan and then a professional translator postedited the output. This English-Catalan test set on the biomedical domain is freely available on request to authors. Details on the test sets are reported in Table \ref{tab:sets}.


\section{Results}

The results of each of the pivotal translation systems
as well as the combined cascaded translation are summarized
in table \ref{baseline-bleu-table}, which shows the high
quality of the translations of the attentional architecture
from \cite{vaswani:2017}.

The English-to-Spanish translation obtains a BLEU score
of 46.55 in the test set of the WMT Biomedical test set while
the Spanish-to-Catalan translation obtains a BLEU score of 86.89
in the El Periódico test set. The cascaded translation
achives a BLEU score of 41.38 in the translated WMT Biometical
test set. 

\begin{table}[H]
\caption{\label{baseline-bleu-table} BLEU results.}
\label{tab:sets}
\small
\centering
\medskip
\begin{tabular}{l cc}
Language & System & BLEU\\ \hline
EN2ES & Direct & 46.55\\
ES2CA & Direct & 86.89\\
EN2CA & Cascade & 41.38\\ \cline{2-3}
\end{tabular}
\end{table}

All BLEU scores are case-sensitive and where obtained
with script \texttt{t2t-bleu} from the tensor2tensor framework,
whose results are equivalent to those from \texttt{mteval-v14.pl}
from the Moses package.


In order to illustrate the quality of the cascaded
translations quality, some sample translations are
shown in table \ref{tab:samples}.




\section{Conclusion}


This paper describes the data resources and architectures to build an English-Catalan neural MT system in the medical domain without English-Catalan parallel resources. This descriptive paper provides details on latest architectures in neural MT based on attention mechanisms and one standard pivot architecture that has been used with the statistical approach. The paper reports results on the baseline system of the cascade approach with the latest neural MT architecture of the Transformer.

Further experiments are required to fully characterize the
potential of pivoting approaches. One of the future lines of
research is to apply the pseudo-corpus approach, which consists
in training a third translation system on a synthetic corpus
created by means of the pivotal ones.
A second future line of research is the use of recently proposed
unsupervised machine translation approaches
\cite{lample18,artetxe18},
which do not require large amount of parallel data.

\section{Acknowledgements}

This work is supported by the Spanish Ministerio de Econom\'ia y Competitividad and European Regional Development Fund, through the postdoctoral senior grant \textit{Ram\'on y Cajal}, the contract TEC2015-69266-P (MINECO/FEDER, UE) and the contract PCIN-2017-079 (AEI/MINECO). This work is also supported in 
part by an Industrial PhD Grant from 
the Catalan Agency for Management of University and Research Grants (AGAUR) and by the United Language Group (ULG).

\clearpage

\section{Bibliographical References}
\label{main:ref}

\bibliographystyle{lrec}
\bibliography{encabio}

\end{document}